\documentclass[letterpaper, 10 pt, conference]{ieeeconf}  % Comment this line out if you need a4paper

\IEEEoverridecommandlockouts                              % This command is only needed if 
\usepackage{graphicx}
\usepackage[font=small,labelfont=bf]{caption}
\usepackage{amsmath} % assumes amsmath package installed
\usepackage{amssymb}  % assumes amsmath package installed
\usepackage{booktabs}
\usepackage{gensymb}
\usepackage{cite}
\usepackage{multirow}
\usepackage{xspace}

\title{\LARGE \bf
Objects Matter: Learning Object Relation Graph for Robust Camera Relocalization
}

\author{Chengyu Qiao$^{1}$, Zhiyu Xiang$^{2}$ and Xinglu Wang$^{3}$% <-this % stops a space
\thanks{All authors are with the Dept of Information and Electronic Engineering,
		Zhejiang University, China. Zhiyu Xiang is the corresponding author,
		xiangzy@zju.edu.cn.}%
}

\begin{document}

\maketitle
\thispagestyle{empty}
\pagestyle{empty}

%%%%%%%%%%%%%%%%%%%%%%%%%%%%%%%%%%%%%%%%%%%%%%%%%%%%%%%%%%%%%%%%%%%%%%%%%%%%%%%%
\begin{abstract}

Visual relocalization aims to estimate the pose of a camera from one or more images. In recent years deep learning based pose regression methods have attracted many attentions. They feature predicting the absolute poses without relying on any prior built maps or stored images, making the relocalization very efficient. However, robust relocalization under environments with complex appearance changes and real dynamics remains very challenging. In this paper, we propose to enhance the distinctiveness of the image features by extracting the deep relationship among objects. In particular, we extract objects in the image and construct a deep object relation graph (ORG) to incorporate the semantic connections and relative spatial clues of the objects. We integrate our ORG module into several popular pose regression models. Extensive experiments on various public indoor and outdoor datasets demonstrate that our method improves the performance significantly and outperforms the previous approaches.

\end{abstract}

%%%%%%%%%%%%%%%%%%%%%%%%%%%%%%%%%%%%%%%%%%%%%%%%%%%%%%%%%%%%%%%%%%%%%%%%%%%%%%%%
\section{INTRODUCTION}
\label{sec:intro}

Visual relocalization refers to the task of recovering the absolute pose of a camera from a single or multiple images, which is one of the fundamental tasks in lots of fields such as robotics, autonomous driving, and virtual/augmented reality. Due to the existence of large perspective and appearance changes, robust relocalization under real environment remains a challenging task.

Lots of effects have been made to deal with this problem. Structure-based algorithms \cite{sattler2016efficient,taira2018inloc,sarlin2021back,sarlin2019coarse} build a prior 3D map for the environment and relocalize the camera by establishing 2D-3D correspondences. They can achieve state-of-the-art performance, but require large memory space to store the map and can only work for known environments. Image retrieve based methods \cite{arandjelovic2016netvlad,sattler2016large,torii201524} search for the most similar images in the stored database and use the pose of the matched image as the recoverd position. They can further be optimized by estimating the relative pose between query and the matched images. Besides storing an image database, image retrieve based methods tend to fail when the database does not cover the novel pose of the query image.

In recent years, a series of Absolute Pose Regression (APR) approaches \cite{kendall2015posenet,kendall2016modelling,kendall2017geometric,melekhov2017image,cai2019hybrid,bui2019adversarial,wang2019discriminative,wu2017delving} are proposed to directly predict the camera pose from an image, without relying on any prior built maps or stored image database. The pose of the camera can also be further improved by integrating constraints from multiple neighboring images with the help of visual odometry (VO) \cite{brahmbhatt2017mapnet,valada2018deep,xue2019local,radwan2018vlocnet++,clark2017vidloc}, Long-Short Term Memory Networks (LSTMs) \cite{xue2019local,clark2017vidloc,walch2017image} or Graph Neural Network (GNNs) \cite{xue2020learning}. Compared to the structure based or retrieval based counterparts, APR approaches learn the entire localization pipeline and are very efficient during runtime. They are an order of magnitude faster and require only a small memory footprint. However, current APR methods are also an order of magnitude less accurate than the structure-based approaches.

\begin{figure}[t]
	\centering
	\includegraphics[width=1.0\linewidth]{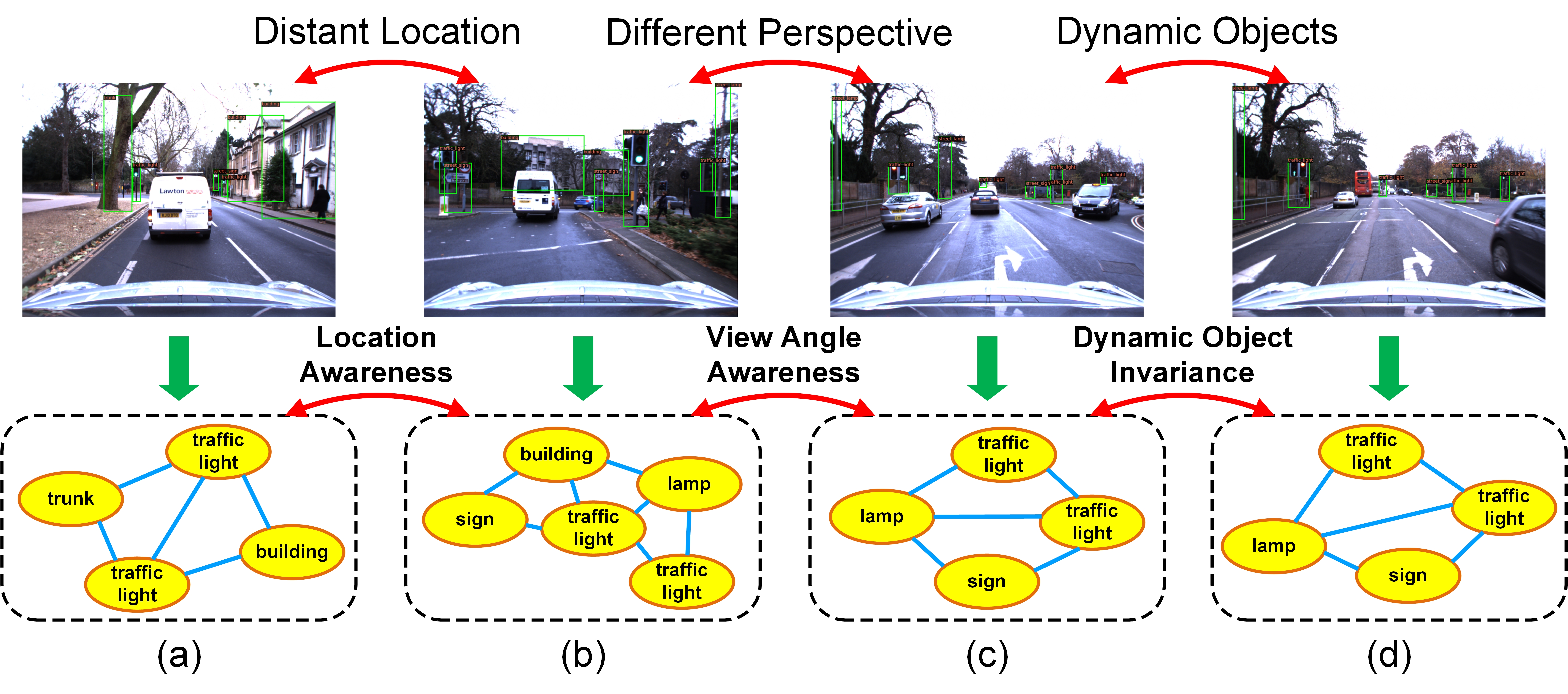}
	\caption{Illustration of our proposed ORG features in different situations. Each image corresponds to a specific ORG that contains semantic connections and relative spatial cues for objects. For image pairs from distant locations shown in (a) and (b), the large difference of the ORG features makes it easy to distinguish poses. For (b) and (c), the vehicle heads towards the same intersection from different routes, resulting in large perspective changes between images. In this situation, the co-occurrence relationship and the scale of objects in ORG help to clarify the current camera pose. At last, the relation features extracted from objects with certain categories are immune to the dynamic objects, leading to more robustness to dynamic environments.}
	\label{fig:teaser}
\end{figure}

In this paper, we focus on improving the robustness of APR methods. No matter one or more images are used, current APR works mainly rely on convolutional network backbones similar to the ones used in image classification task to extract features of the image. However, different from the image classification problem where the pose and scale invariant features are desired, the ideal feature for the relocalization task should be pose and scale awareness while keeping invariance to the disturbances like illumination changes and dynamic objects.

In this paper, we propose a novel Object Relation Graph (ORG) to extract object relation feature and strengthen the inner representation of an image for relocalization. The objects in the image and the relationships among them can be important clues for the recovery of the camera pose in complex scenes, in that they can provide semantic stability and pose/scale awareness, as shown in Fig~\ref{fig:teaser}. In most of the scenes, objects and the co-occurrence relationships among them are relatively stable. They do not change with the imaging projection, and can provide robust global localization of the camera on semantic space. Meanwhile, with the change of local position and view angle, the size of objects and the image distances among them change accordingly, which can facilitate local localization. When dealing with dynamic environments, certain dynamic object categories, e.g., vehicles and pedestrians, can be safely removed from the object relation graph, providing enhanced robustness to such scenes. We extract object features by employing Graph Neural Networks (GNNs). Specifically, we first detect objects in the image and obtain the primitive object features that contain spatial and semantic information. Then, the features are fed into a deep GNN where each node aggregates information from neighboring nodes with dynamic edge connections. To make the network more adaptive to the visual relocalization task, we redesign the nodes, edges, and embedded functions of GNNs. Finally, we integrate our Object Relation Graph model into PoseNet \cite{kendall2017geometric} and MapNet \cite{brahmbhatt2017mapnet}, which are the popular single-frame and multi-frame based APR frameworks respectively. Both of the resulting frameworks are differentiable and can be trained end-to-end.

In summary, our contributions can be summarized as:
\begin{itemize}
	\item We explore the role of object relationship for the task of absolute pose regression. The object features with stable semantic co-occurrence and pose/scale awareness can provide vital clues to the task. To the best of our knowledge, we are the first to tackle this task by enhancing the visual representation with explicitly learned object relation features.
	\item We propose a novel object relation graph (ORG) for object feature extraction, which utilizes deep multi-layer GNNs with dynamic edges to exploit semantic and spatial position relationships among objects.
	\item We integrate our ORG module into popular APR networks and carry out extensive experiments on various public datasets. The results show that our method has a significant improvement upon the baselines and outperforms previous absolute pose regression approaches.
\end{itemize}

\section{Related Work}
\label{sec:related}

\subsection{Visual Relocalization}

Current Visual relocalization methods can be divided into three families according to their algorithmic characteristics, e.g., structure based methods, retrieval based methods and APR methods.

\textbf{Structure based methods} \cite{sattler2016efficient,brachmann2017dsac,brachmann2018learning,taira2018inloc,sarlin2019coarse,sarlin2021back} are composed of a complex pipeline starting from building a prior 3D map for the scene, then establishing 2D-3D correspondences for the query image and finally estimating the camera pose using PnP with outlier rejection. The correspondences are usually established by descriptor matching \cite{sattler2016efficient,sarlin2019coarse,sarlin2021back}. Recently, DSAC \cite{brachmann2017dsac} and the follow-up DSAC++ \cite{brachmann2018learning} propose to train a CNN architecture to regress the 3D coordinates for pixels in the image, thus the time-costing descriptor based 2D-3D matching can be avoided. Despite their high accuracy in small and medium scale environment, structure based methods are memory and computing intensive, which restricts its application in large scale environments.

\textbf{Retrieval based methods} \cite{arandjelovic2016netvlad,sattler2016large,torii201524} recover absolute camera poses by either using the pose of the most similar image found in the database or further estimating the relative pose between the query image and the matched one. They utilize image-level descriptors for effcient and scalable retrieval. Such approaches require searching through large database and tend to fail when the database does not cover the novel pose of the query image.

\textbf{Absolute pose regression} has become a popular method for relocalization in recent years. It does not require any prior map or stored image database during runtime. PoseNet \cite{kendall2015posenet} and its variants \cite{kendall2016modelling,kendall2017geometric,melekhov2017image,walch2017image,cai2019hybrid,wang2019discriminative,wu2017delving,bui2019adversarial} regress camera pose with a single image as input. In order to take advantage of sequential images, VidLoc \cite{clark2017vidloc} uses an LSTM to learn the temporal relationship between consecutive frames. MapNet \cite{brahmbhatt2017mapnet} further applies the pose constraints between neighboring images and integrates visual odometry to improve localization accuracy. \cite{sattler2019understanding} summarizes the state-of-the-art APR methods and points out the limitation that APR methods tend to do interpolation between seen camera poses in the training set. Recently, LSG \cite{xue2019local} and GRNet \cite{xue2020learning} use LSTMs and GNNs respectively to model the smoothness of motion within the image sequences. NeuralR-Pose \cite{zhu2021learning} proposes to learn neural representations for camera poses and compute the pose loss in the learned neural space. Currently, the APR methods are still far less accurate than structure-based methods, with tens of meters of mean localization error in large environments. But it is much more memory friendly and computing efficient.

\subsection{Object Related Methods}

There exists some approaches that utilize objects for localization. As an additional constraint, \cite{salas2013slam++,li2019semantic} incorporate the detected objects into the traditional Simultaneously Localization and Mapping (SLAM) pipeline to improve the robustness in the case of viewpoint changes.  \cite{liu2019global} extracts the object-level semantics to build a global semantic map, and estimates the pose of the query image within the global map by aligning the associated objects. \cite{weinzaepfel2019visual,ardeshir2014gis} find 2D-2D dense matches between the detected objects and the objects in the database to localize the query image. Our methods differs from them by extracting object relationship and integrate it into APR approaches.

Object relationship has been employed in a few computer vision tasks such as scene graph generation \cite{dai2017detecting}. It encodes the interplay between the object instances to achieve high-level image understanding to facilitate applications such as image capturing \cite{wang2020improving} and manipulation \cite{dhamo2020semantic}. However, such kind of high level relationship is too general and lacks of discrimination for the low-level camera relocalization task at hand.

\section{Method}
\label{sec:method}

\subsection{Acquiring Primitive Object Feature}

In order to take full advantage of the information of objects and capture the semantic and spatial relationships among objects, we first need to detect the objects of interest in the image. We leverage the Fully Convolutional One-Stage (FCOS) \cite{tian2019fcos} which is a real-time object detection network to conduct object detection. Given an RGB image as input, FCOS detects desired objects in the image and output their bounding box positions and labels. For $n$ detected objects $o_i$ where $i\in \left\{1,2,...,n\right\}$, we parameterize the detection as a tuple of $o=\left\{x,y,w,h,C\right\}$, where $(x,y)$ denotes the pixel coordinates of the object center, $(w,h)$ denotes the object size with its width and height, and $C$ is the object category ID. Note that the objects of interest are different for indoor and outdoor scenes due to the huge difference between the two scenes. Objects that frequently appear in indoor scenes such as TVs or tables are usually static. However, objects like vehicles and people that are usually more interested by object detection methods for outdoor scenes are dynamic and may have an adverse impact on the relocalization task. Thus, we train different object detectors for indoor and outdoor scenes and only detect those static objects such as tree trunks, buildings and traffic lights for outdoor scenes. The detection result $o_i$ is fed into an MLP (Multi-Layer Perceptron) to obtain the primitive object feature $x_i$, which contains the object-level information and will be used as the node feature in the graph later.

\subsection{Graph Definition} 

For the detected objects and corresponding features, we model them using our object relation graph (ORG). We first define the graph as $\mathcal{G} = \{\mathcal{V}, \mathcal{E}\}$ with $\mathcal{V}=\{v_1,...,v_n\}$ and $\mathcal{E} \subseteq \mathcal{V} \times \mathcal{V} $ representing the nodes and edges respectively. Each detected object is regarded as a node $v_i$ which contains the corresponding object feature $x_i$. Each edge $e_{ij} = (v_i,v_j)$ denotes the link among different objects.

As each edge determines whether two objects are connected and allows them to exchange information, we need to define the edge connection in the initial stage. One direct way is to densely connect each pair of nodes, which will result in a fully-connected graph. However, it can lead to redundancy and expensive computational cost, especially when there are many objects detected in the image. In addition, full connection is not beneficial to capture the relationships among those objects that are really related. Therefore, we determine the existence of edges based on the distance between node features. Specifically, each node is connected to k nearest neighbors who have the smallest Euclidean distances to the current node in feature space.

\subsection{Learning Object Relation Graph}

\begin{figure}[t]
	\centering
	\includegraphics[width=0.8\linewidth]{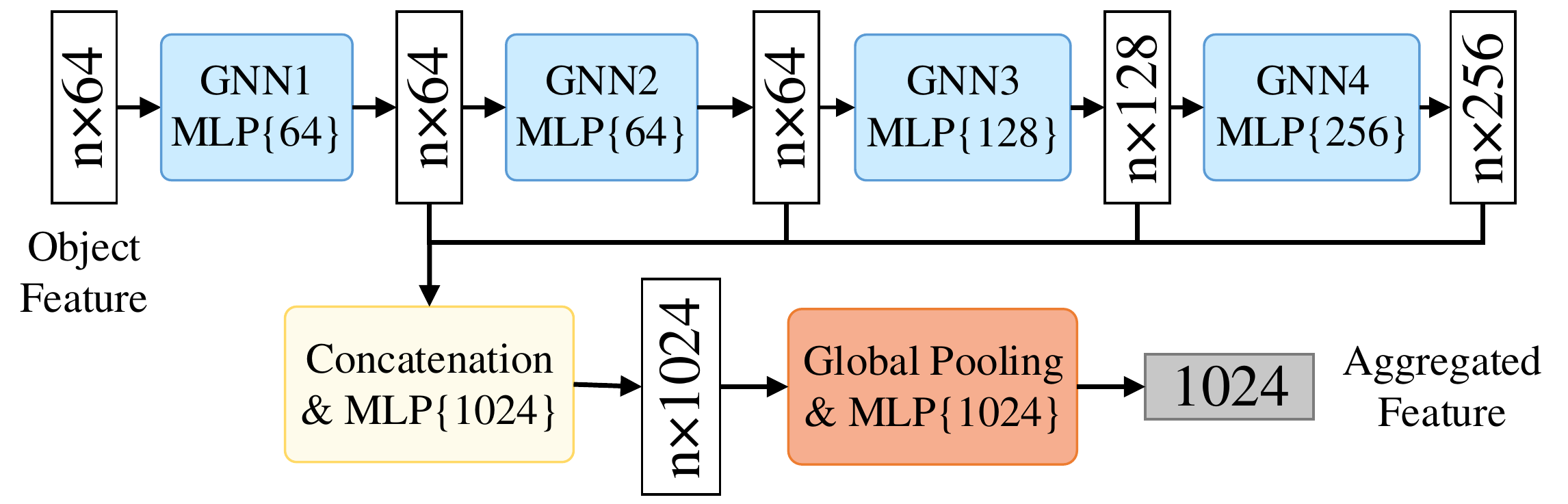}
	\caption{The architecture of the Object Relation Graph. The detection features of $n$ objects are fed to four consecutive GNN layers for information exchange. Fused features from the four GNN layers are concatenated along the channel dimension, and the aggregated object relation features are obtained through global pooling. ``MLP'' stands for multi-layer perceptron, and numbers in brackets are layer sizes.}
	\label{fig:structure}
\end{figure}

Given the nodes and edges of a graph, our goal is to learn a better object relationship through transferring information between nodes. Next, we introduce the method of node feature update in our graph neural network.

\textbf{Neighboring Differential Feature Generation.~}
In order to update the features of nodes, we first need to capture the relationship between nodes. For node $v_i$, we first calculate the feature distances between node $v_i$ and all of the rest nodes $v_j$ in graph $\mathcal{G}$. The distance $d_{ij}$ is defined as:
\begin{equation}
d_{ij} = \left\Vert x_i - x_j \right\Vert_2
\end{equation}
Then we take $k$ nodes $v_{j_{i1}},\hdots,v_{j_{ik}}$ with the smallest distance to $v_i$, and establish $k$ directed edges for node $v_i$ in the form of $(i,j_{i1}),\hdots,(i,j_{ik})$. In this way, we construct $\mathcal{G}$ as a directed $k$-nearest neighbor ($k$-NN) graph. Given the neighboring node $v_j$, the neighboring differential feature $n_{ij}$ for node $v_i$ are computed as $n_{ij} = x_j - x_i$, which will be used in feature aggregation. 

\textbf{Node Feature Aggregation.~}
Inspired by \cite{qi2017pointnet} who uses a symmetric function to aggregate point information, the node feature aggregation function is defined as:
\begin{equation}
x_i' = \underset{j:(i,j)\in \mathcal{E}}{\mbox{MAX}}\left\{f_u(x_i,n_{ij})\right\}
\label{eq:ag}
\end{equation}
where $f_u$ is a fully connected layer followed by a BatchNorm layer and a ReLU non-linearity, and $\mbox{MAX}$ is a vector max operator which takes $k$ features as input and returns a new element-wise maximum vector. $x_i'\in \mathbb{R}^{d'}$ is the aggregated feature of node $v_i$, and $d'$ is the dimension of the resulting feature. $x_i$ and $n_{ij}$ are concatenated along the channel dimension before feeding into $f_u$. The resulting feature fuses object attributes $x_i$ (contributed by the object positions and categories) and its local neighbor information $n_{ij}$ to obtain a higher level of object representation..

\subsection{Dynamic Graph Update}

\begin{figure*}[t]
	\centering
	\includegraphics[width=1.0\linewidth]{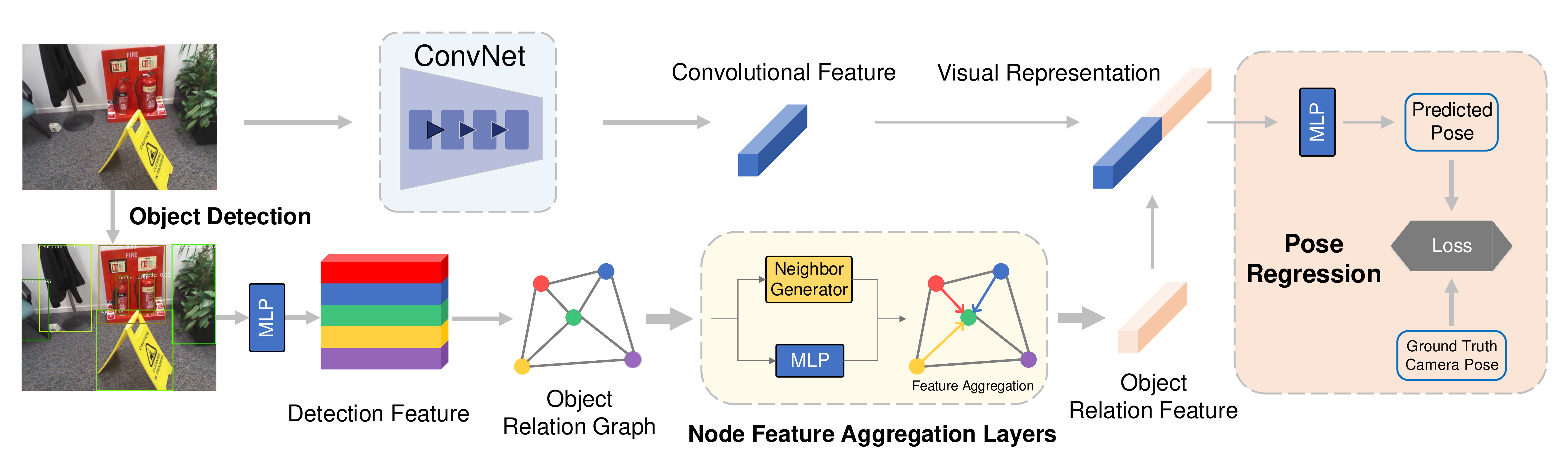}
	\caption{Overview of the implemented framework. Given the input image, the upper branch is the baseline network and the lower branch is the pipeline which includes the ORG module. The convolutional features and the object relation feature are concatenated before feeding into the pose regression head.}
	\label{fig:network}
\end{figure*}

After applying neighbor feature generation and node feature aggregation once, the features of the nodes in the graph are updated. This can be seen as one layer in the graph network. Similar to applying the convolutional layer multiple times to obtain a larger receptive field, we apply the GNN layer multiple times so that nodes can obtain information from more other nodes. Different from the traditional GNNs in which edges are pre-defined and fixed, our ORG network dynamically defines edges at each layer according to the updated node features, resulting in a dynamic graph update. Let $X^l = \left\{x_1^l,\hdots,x_n^l\right\}\subseteq \mathbb{R}^{d_l}$ be the output of the $l$-th layer, the updated node feature of the $(l+1)$-st layer is
\begin{equation}
x_i^{l+1} = \underset{j:(i,j)\in \mathcal{E}^l}{\mbox{MAX}}\left\{f_u^{l+1}(x_i^l,n_{ij}^l)\right\}
\end{equation}
where $f_u^{l+1}:\mathbb{R}^{2d_l} \rightarrow \mathbb{R}^{d_{l+1}}$ is the corresponding MLP layer similar to Eq.~\ref{eq:ag}.

As shown in Fig~\ref{fig:structure}, a four-layer GNN structure is designed in this paper to extract multi-level object features. At each GNN layer, $k$ edges are dynamically constructed according to the updated features. The multi-level object features from four GNN layers are then concatenated and fed into a fully connected layer to obtain a fused feature $x^a$ with dimension $d_a$. Finally a symmetry function $f_g$ is used to aggregate the information from all objects, as shown in Eq.~\ref{eq:fg}.
\begin{equation}
x_g = f_g\left(\underset{i=1,\hdots,n}{\mbox{MAX}}\left\{x_i^a\right\},\underset{j=1,\hdots,n}{\mbox{AVG}}\left\{x_j^a\right\}\right)
\label{eq:fg}
\end{equation}
where $x_g\subseteq \mathbb{R}^{d_g}$ is the final object relation feature for the image, $\mbox{AVG}$ is a vector average operator, which takes $k$ features as input and returns a new element-wise averaged vector, $f_g:\mathbb{R}^{2d_a} \rightarrow R^{d_g}$. In this step, the fused features are passed through a GMP (global max pooling) layer and a GAP (global average pooling) layer respectively for downsampling and then concatenated along the channel dimension, and finally fed into a fully connected layer to obtain a vector representing the object relation feature for the entire image.

\subsection{Relocalization Framework}

Our object relation feature can be regarded as a valuable and complementary feature to the existing convolution based features in the relocalization task. Therefore, we embed our ORG module into the existing APR method. The resulting relocalization framework is shown in Fig~\ref{fig:network}. We select two APR methods respectively as our baselines to construct the framework. The first baseline is PoseNet \cite{kendall2017geometric}, which is a classical APR network with only one image as input. As pointed out in \cite{sattler2019understanding}, most of the state-of-the-art APR methods are designed in accordance with PoseNet \cite{kendall2017geometric}. For the methods using sequential images as input, MapNet \cite{brahmbhatt2017mapnet} is a good choice. It follows the network structure of PoseNet and adds an additional loss term during training to further embed the multi-frame constraints. The resulting frameworks are called ORGPoseNet and ORGMapNet respectively. Note that the main focus of this work is to explore the effect of introducing object relation features, therefore we do not include other improvement techniques (e.g., including attention mechanisms or exchanging information between frames via LSTMs or GNNs), as we consider them as orthogonal directions to improve the pose representations. As shown in Fig~\ref{fig:network}, the object relation feature $x_g$ from the ORG module and the convolutional feature $x_c$ from the baseline network are concatenated together before feeding into the pose regression head.

The pose regression head of the framework contains nothing but a fully connected layer followed by a dropout layer and two fully connected layers to predict the translation $\mathbf{t}\in \mathbb{R}^3 $ and rotation $\mathbf{r}\in \mathbb{R}^3 $ of the image respectively. The predicted rotation term follows the form of MapNet, which is the logarithm of a unit quaternion.

\subsection{Loss Function}

As in \cite{kendall2017geometric,brahmbhatt2017mapnet}, the loss for ORGPoseNet is defined as:
\begin{equation}
L_{single} = \sum_{i=1}^{N}d(\mathbf{p}_i,\mathbf{p}_i^*)
\end{equation}
where $\mathbf{p}_i = (\mathbf{t}_i,\mathbf{r}_i)$ and $\mathbf{p}_i^* = (\mathbf{t}_i^*,\mathbf{r}_i^*)$ are the predicted and the ground truth absolute poses, respectively. $d(\cdot)$ is a distance function defined as:
\begin{equation}
d(\mathbf{p}_i,\mathbf{p}_i^*) = \left\Vert\mathbf{t}_i - \mathbf{t}_i^*\right\Vert_1e^{-\beta} + \beta + \left\Vert\mathbf{r}_i - \mathbf{r}_i^*\right\Vert_1e^{-\gamma} + \gamma
\label{eq:df}
\end{equation}
where $\beta$ and $\gamma$ are learnable parameters used to balance the translation and rotation loss, and they are optimized jointly with other parameters in the neural network during the training process.

For ORGMapNet, the loss function also takes the relative pose loss into account. It is defined as \cite{brahmbhatt2017mapnet}:
\begin{equation}
L_{frame} = \sum_{i=1}^{N}d(\mathbf{p}_i,\mathbf{p}_i^*) + \sum_{i,j=1,i\neq j}^{N}d(\mathbf{v}_{ij},\mathbf{v}_{ij}^*)
\end{equation}
where $\mathbf{v}_{ij} = (\mathbf{t}_i - \mathbf{t}_j,\mathbf{r}_i - \mathbf{r}_j)$ is the relative pose between the predicted pose $\mathbf{p}_i$ for the image $I_i$ and the predicted pose $\mathbf{p}_j$ for the image $I_j$. $(I_i,I_j)$ is an image pair in each tuple of $s$ images, and these images are sampled at intervals of $k_f$ frames in the training set.

\section{Experiments}
\label{sec:experiments}

\subsection{Implementation Details}

Following \cite{brahmbhatt2017mapnet}, we employ ResNet34 \cite{he2016deep} pretrained on ImageNet \cite{deng2009imagenet} to extract convolution features with dimension 1024. In the ORG branch, the graph is constructed using $k=5$ nearest neighbors. The dimension of the final output vector $x_g$ representing the relationship between objects is also 1024. For pose regression head, we use a fully connected layer with hidden feature dimension of $2048 \Rightarrow 512$, followed by two fully connected layers with dimension $512 \Rightarrow 3$ for translation and rotation respectively.

We use PyTorch \cite{paszke2019pytorch} to implement our model on a single GTX 2080Ti GPU. As in \cite{kendall2017geometric,brahmbhatt2017mapnet}, all input RGB images are resized with the height to 256 and normalized by pixel mean subtraction and standard deviation division. We set $\beta$ and $\gamma$ to 0.0 and -3.0 for initialization, respectively. For ORGMapNet, the size $s$ and sampling interval $k_f$ of the input image sequence are set to 3 and 10, respectively, as in \cite{brahmbhatt2017mapnet}. We use the Adam optimizer \cite{kingma2014adam} with a weight decay of $5 \times 10^{-4}$. The learning rate of the ResNet34 network is set to $1 \times 10^{-4}$ throughout the training process, while the learning rate of other parts is $1 \times 10^{-3}$ at the beginning and divided by 10 at the 300-th epoch. All models are trained for 600 epochs with batch size of 64 and 20 for ORGPoseNet and ORGMapNet, respectively. The resulting network can run at a rate of 51.6 FPS with a single GPU during inference.

\subsection{Settings}

\textbf{Datasets.~}
We evaluate our algorithm on three well-known public datasets with different environmental scales, including 7-Scenes \cite{shotton2013scene}, RIO10 \cite{wald2020beyond} and Oxford RobotCar \cite{maddern20171}.

The 7-Scenes dataset \cite{shotton2013scene} is an indoor dataset containing seven different scenes recorded by a Kinect RGB-D sensor. The groundtruth camera poses are obtained by applying the KinectFusion system. Each scene contains a specific room in an office building, in which different trajectories are recorded to obtain multiple sequences. The scenes contain many textureless regions, which is not easy for visual relocalization.

The RIO10 dataset \cite{wald2020beyond} contains ten RGB-D image sequences of indoor scenes captured by a mobile phone. Each scene is scanned multiple times over a period of up to one year. The groundtruth camera poses are obtained by an offline bundle adjustment framework. In addition to RGB images and depth images, the dataset also provides semantic maps for reference. Dynamic objects, changes in long-term spans, blurring and illumination changes captured by the mobile phone make this dataset very challenging for relocalization task.

The Oxford RobotCar dataset \cite{maddern20171} contains over 100 repetitions of a continuous route through Oxford over a year. The images are collected from a camera mounted on a vehicle. Compared with the indoor datasets, this dataset contains much longer trajectories and larger areas, which brings greater challenges to the relocalization methods due to the complex dynamic outdoor environment. In addition, the dataset images are collected under different lighting and weather conditions, which makes it an ideal choice for evaluating the method in real outdoor scenes.

\textbf{Evaluation Metrics.~}
We use absolute pose errors for relocalization evaluation, as in \cite{kendall2015posenet}. In addition, in the RIO10 dataset, we further adopt their proposed new metrics such as Dense Correspondence Re-Projection Error (DCRE), Outlier and Score. DCRE is calculated as the average magnitude of the 2D correspondence displacement, normalized by the image diagonal. The displacement is calculated according to the 2D projection of dense 3D points rendered from an underlying 3D model using ground truth and predicted camera poses. DCRE(0.05) and DCRE(0.15) represent the proportion of frames with DCRE lower than 0.05 and 0.15 respectively, while Outlier(0.5) is the proportion of frames whose DCRE is above 0.5. The Score metric is defined as the sum of 1 and the difference between DCRE(0.15) and Outlier(0.5) to represent the general quality of the result.

\subsection{Experiments on the 7-Scenes Dataset}

\begin{table*}[t]
	%\footnotesize
	\small
	\centering
	\renewcommand{\arraystretch}{1.5}
	\resizebox{\textwidth}{!}{
	\begin{tabular}{lcccccccc}
		\toprule
		\multirow{2}{*}{} & \multicolumn{8}{c}{Sequence} \\
		Method & Chess & Fire & Heads & Office & Pumpkin & Kitchen & Stairs & Average \\
		\midrule
		PoseNet17 \cite{kendall2017geometric}       & 0.14m, 4.50\degree & 0.27m, 11.80\degree & 0.18m, 12.10\degree & 0.20m, 5.77\degree & 0.25m, 4.82\degree & 0.24m, 5.52\degree & 0.37m, 10.60\degree & 0.24m, 7.87\degree \\
		PoseNet + log q \cite{brahmbhatt2017mapnet} & 0.11m, 4.29\degree & 0.27m, 12.13\degree & 0.19m, \textbf{12.15\degree} & 0.19m, 6.35\degree & 0.22m, 5.05\degree & 0.25m, 5.27\degree & 0.30m, 11.29\degree & 0.22m, 8.07\degree \\
		PoseNet + log q (*)                         & 0.17m, 4.96\degree & 0.36m, 11.22\degree & 0.20m, 13.35\degree & 0.23m, 7.05\degree & 0.26m, 5.87\degree & 0.29m, 6.10\degree & 0.36m, 10.18\degree & 0.27m, 8.39\degree \\
		NeuralR-Pose \cite{zhu2021learning}         & 0.12m, 4.83\degree & 0.27m, \textbf{8.91\degree} & 0.16m, 12.84\degree & 0.19m, 6.64\degree & 0.22m, 5.45\degree & 0.24m, 6.10\degree & 0.29m, 10.70\degree & 0.21m, 7.92\degree \\
		ORGPoseNet (Ours)                           & 0.10m, 3.29\degree & 0.33m, 11.02\degree & \textbf{0.15m}, 13.34\degree & 0.19m, 5.91\degree & 0.20m, 5.42\degree & 0.24m, 5.71\degree & 0.27m, 10.63\degree & 0.21m, 7.90\degree \\
		MapNet \cite{brahmbhatt2017mapnet}          & \textbf{0.08m}, \textbf{3.25\degree} & 0.27m, 11.69\degree & 0.18m, 13.25\degree & \textbf{0.17m}, 5.15\degree & 0.22m, \textbf{4.02\degree} & 0.23m, \textbf{4.93\degree} & 0.30m, 12.08\degree & 0.21m, 7.77\degree \\
		ORGMapNet (Ours)                            & 0.09m, 3.60\degree & \textbf{0.26m}, 9.49\degree & \textbf{0.15m}, 12.81\degree & 0.20m, \textbf{4.96\degree} & \textbf{0.18m}, 5.04\degree & \textbf{0.22m}, 5.68\degree & \textbf{0.27m}, \textbf{9.54\degree} & \textbf{0.20m}, \textbf{7.30\degree} \\
		\bottomrule
	\end{tabular}
	}
	\caption{Pose errors of the image-based methods and the sequence-based methods on the 7-Scenes dataset. The column PoseNet + log q (*) are the results of running the code provided by \cite{brahmbhatt2017mapnet} as \cite{zhu2021learning}. Following the convention, the median prediction errors are reported here. The best results are highlighted.}
	\label{tab:result_7scenes}
\end{table*}

We compare our implemented models respectively with the two baseline methods as well as some state-of-the-art methods such as NeuralR-Pose \cite{zhu2021learning}. For the PoseNet \cite{kendall2017geometric}, besides baseline using log quaternions, the results of its variation versions are also listed. The quantitative results are shown in Table~\ref{tab:result_7scenes}. It can be seen that our model outperforms the baselines in both translation and rotation prediction, especially in repetitive and weak texture scenes, e.g., Stairs and Heads. This is because by constructing the object relation graph, our method learns a more robust image representation for changes in illumination and viewing angles, thus contributing to pose estimation. It is not surprising that ORGMapNet performs better than ORGPoseNet, since the former takes the sequential image as input and provides more constraints on the resulting pose. More details can be found in the Supplementary.

\subsection{Experiments on the RIO10 Dataset}

Containing the indoor images recorded over a year, the RIO10 dataset \cite{wald2020beyond} is suitable for testing the method on long-term relocalization task. We compare our method with the baselines as well as some representative image retrieval methods \cite{sarlin2019coarse,arandjelovic2016netvlad,torii201524}. As shown in Table~\ref{tab:result_rio10}, our ORGMapNet outperforms other methods on all of the metrics. It shows the superiority and large potential of the learning based absolute pose regression methods over the image retrieval ones. When compared with the corresponding baselines, our methods significantly improve the accuracy of pose prediction and reduce the outlier ratio. Especially for ORGMapNet, Outlier(0.5) error is brought down by 17.05\% and Score is raised by about 16\% upon the MapNet baseline. This should thank to the object relation graph which captures the location and semantic relationships of objects in the scene. The enhanced object relation features improve the robustness to appearance changes caused by long-term environmental variation.

\begin{table*}[t]
	%\footnotesize
	\centering
	\resizebox{\textwidth}{!}{
	\begin{tabular}{lccccc}
		\toprule
		Method & Score $\uparrow$ & DCRE(0.05) $\uparrow$ & DCRE(0.15) $\uparrow$ & ($\Delta t$,$\Delta \theta$) $\downarrow$ & Outlier(0.5) $\downarrow$ \\
		\midrule
		HFNet \cite{sarlin2019coarse}            & 0.384 & 0.057 & 0.098 & (1.56m, 72.33\degree) & 0.714 \\
		NetVLAD \cite{arandjelovic2016netvlad}   & 0.673 & 0.006 & 0.125 & (0.93m, 31.44\degree) & 0.452 \\
		DenseVLAD \cite{torii201524}             & 0.604 & 0.008 & 0.124 & (0.98m, 32.26\degree) & 0.520 \\
		PoseNet \cite{brahmbhatt2017mapnet}      & 0.624 & 0.012 & 0.149 & (0.89m, 41.89\degree) & 0.525 \\
		ORGPoseNet (Ours)                        & 0.662 & 0.010 & 0.153 & (0.86m, 37.23\degree) & 0.491 \\
		MapNet \cite{brahmbhatt2017mapnet}       & 0.682 & 0.030 & 0.192 & (0.84m, 37.43\degree) & 0.510 \\
		ORGMapNet (Ours)                         & \textbf{0.791} & \textbf{0.064} & \textbf{0.214} & (\textbf{0.80m}, \textbf{30.83\degree}) & \textbf{0.423} \\
		\bottomrule
	\end{tabular}
	}
	\caption{Comparison on the validation split of the RIO10 dataset. The best results are highlighted.}
	\label{tab:result_rio10}
\end{table*}

\subsection{Experiments on the RobotCar Dataset}

We evaluate our method on the Oxford RobotCar dataset which consists of challenging $9562m$ long FULL sequences covering an area of $1.2\times10^6m^2$. All models are trained with two sequences captured on the FULL route under overcast weather, while the test sequences are captured in various weather conditions or long time spans. We report both median and mean pose errors on all of the test sequences, median errors focus on reflecting the general accuracy of the model, while mean errors tend to represent more on the proportion of outliers.

Table~\ref{tab:result_robotcar} shows the quantitative results comparing with the baselines and other state-of-the-art methods. Firstly, our methods significantly improve the performance upon the baselines, in the sense that the corresponding average median translation errors are both reduced by about 66\%. Secondly, in FULL3-5 sequences with different weather and long time spans, we can observe the degraded performance for all models. It reveals the great difficulties of the relocalization task posed by these challenging conditions. In the meantime, compared with the baselines, our methods achieve much lower error variances and exhibit more robustness to the changing environment. We believe this is due to the explicit integration of object relation features to the common convolutional visual features. Thirdly, our ORGMapNet also outperforms multi-frame methods like LSG \cite{xue2019local}, AtLoc+ \cite{wang2020atloc} and GRNet \cite{xue2020learning} in the available FULL1-2 sequences. It is worth noting that these methods predict camera pose by aggregating at least 7 images, while our ORGMapNet only takes 3 frames as input. On the other hand, our method is complementary to these multi-frame methods, as it aims to learn a more discriminative feature representation within an image while LSG and GRNet focus on fusing the information of multiple images by LSTM or GNN models. It will be interesting to explore the combination of these ideas in future.

\begin{table*}[t]
	%\small
	%\footnotesize
	\centering
	\renewcommand{\arraystretch}{1.1}
	\resizebox{\textwidth}{!}{
	\begin{tabular}{l|c|cccccc}
		\toprule
		\multirow{3}{*}{Method} & \multirow{3}{*}{ N } & \multicolumn{6}{c}{Scene} \\
		& & FULL1 & FULL2 & FULL3 & FULL4 & FULL5 & Average \\
		& & overcast & overcast & rain & sun & overcast, 1 year & \\
		\midrule
		\multirow{2}{*}{PoseNet \cite{kendall2017geometric}} & \multirow{2}{*}{1} & 107.62m/125.6m, & 101.94m/131.06m, & 81.31m/151.87m, & 161.92m/213.62m, & 115.47m/165.23m, & 113.65m/157.48m, \\
		& & 22.49\degree/27.1\degree & 20.00\degree/26.05\degree & 15.71\degree/35.17\degree & 29.67\degree/42.07\degree & 27.16\degree/34.95\degree & 23.01\degree/33.07\degree \\
		ORGPoseNet                                           & \multirow{2}{*}{1} & 26.58m/53.08m, & 28.33m/72.36m, & 38.45m/160.58m, & 43.74m/127.34m, & 40.71m/129.75m, & 35.56m/108.62m, \\
		(Ours) & & 4.65\degree/10.18\degree & 4.59\degree/16.15\degree & 6.25\degree/33.72\degree & 7.14\degree/24.72\degree & 7.98\degree/24.51\degree & 6.12\degree/21.86\degree \\
		\multirow{2}{*}{MapNet \cite{brahmbhatt2017mapnet}}  & \multirow{2}{*}{3} & 18.21m/41.40m, & 21.16m/59.30m, & 45.51m/167.75m, & 76.51m/208.11m, & 32.44m/107.54m, & 38.77m/116.82m, \\
		& & 6.66\degree/12.50\degree & 6.38\degree/14.81\degree & 10.00\degree/35.56\degree & 13.62\degree/32.15\degree & 9.22\degree/28.98\degree & 9.18\degree/24.82\degree \\
		\multirow{2}{*}{LSG \cite{xue2019local}}             & \multirow{2}{*}{7} & -/31.65m, & -/53.45m, & \multirow{2}{*}{-} & \multirow{2}{*}{-} & \multirow{2}{*}{-} & \multirow{2}{*}{-} \\
		& & -/4.51\degree & -/8.60\degree & & & & \\
		\multirow{2}{*}{AtLoc+ \cite{wang2020atloc}}         & \multirow{2}{*}{3} & \textbf{6.40m}/21.0m, & \textbf{7.00m}/42.6m, & \multirow{2}{*}{-} & \multirow{2}{*}{-} & \multirow{2}{*}{-} & \multirow{2}{*}{-} \\
		& & 1.50\degree/6.15\degree & 1.48\degree/9.95\degree & & & & \\
		\multirow{2}{*}{GRNet \cite{xue2020learning}}        & \multirow{2}{*}{8} & -/17.35m, & -/37.81m, & \multirow{2}{*}{-} & \multirow{2}{*}{-} & \multirow{2}{*}{-} & \multirow{2}{*}{-} \\
		& & -/\textbf{3.47\degree} & -/\textbf{7.55}\degree & & & & \\
		ORGMapNet                                            & \multirow{2}{*}{3} & 8.84m/\textbf{14.41m}, & 8.01m/\textbf{36.61m}, & \textbf{15.15m}/\textbf{103.42m}, & \textbf{14.86m}/\textbf{75.46m}, & \textbf{16.85m}/\textbf{95.03m}, & \textbf{12.74m}/\textbf{64.99m}, \\
		(Ours) & & \textbf{1.13\degree}/3.73\degree & \textbf{1.26\degree}/9.49\degree & \textbf{2.58\degree}/\textbf{20.99\degree} & \textbf{1.79\degree}/\textbf{12.30\degree} & \textbf{2.09\degree}/\textbf{16.85\degree} & \textbf{1.77\degree}/\textbf{12.67\degree} \\
		\bottomrule
	\end{tabular}
	}
	\caption{The camera pose prediction errors on the Oxford RobotCar dataset. ``N'' indicates the number of images input to the network at one time. We report the median / mean translation and rotation errors. The best results are highlighted.}
	\label{tab:result_robotcar}
\end{table*}

\begin{figure}[t]
	\centering
	\begin{minipage}{0.32\linewidth}
		\centering
		\includegraphics[width=1.0\linewidth]{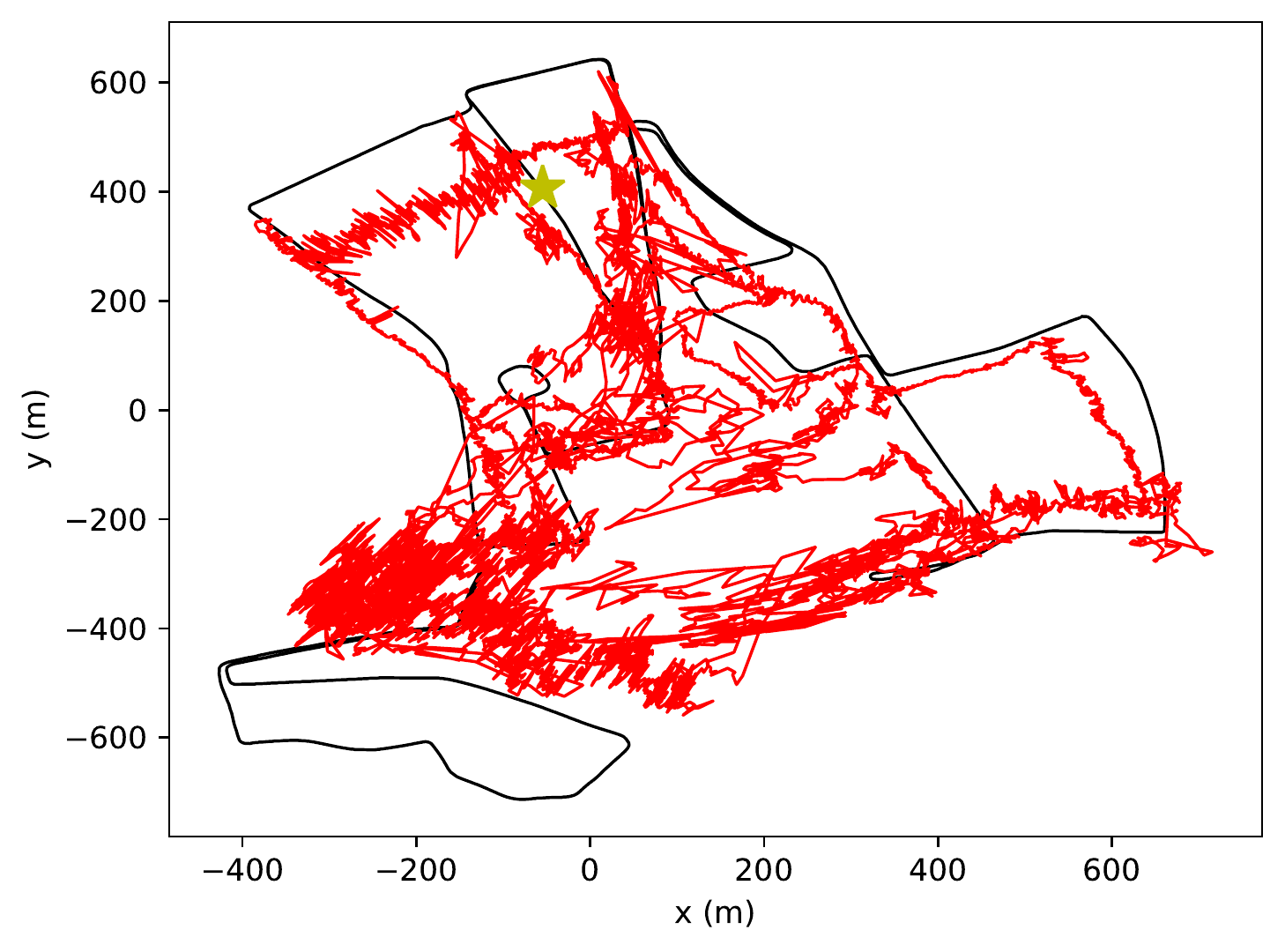}
		PoseNet \cite{kendall2017geometric}
	\end{minipage}
	\begin{minipage}{0.32\linewidth}
		\centering
		\includegraphics[width=1.0\linewidth]{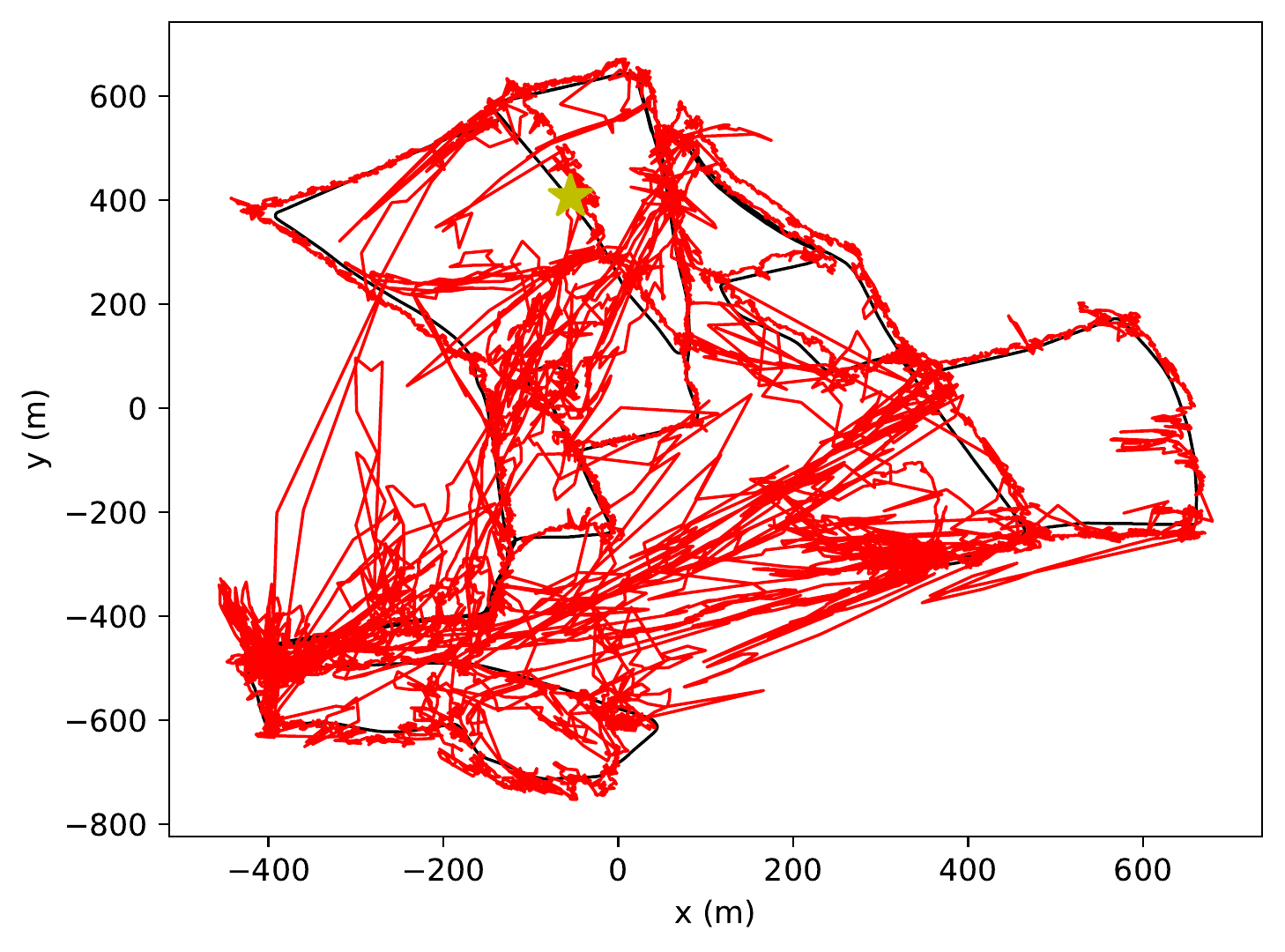}
		ORGPoseNet
	\end{minipage}
	\begin{minipage}{0.32\linewidth}
		\centering
		\includegraphics[width=1.0\linewidth]{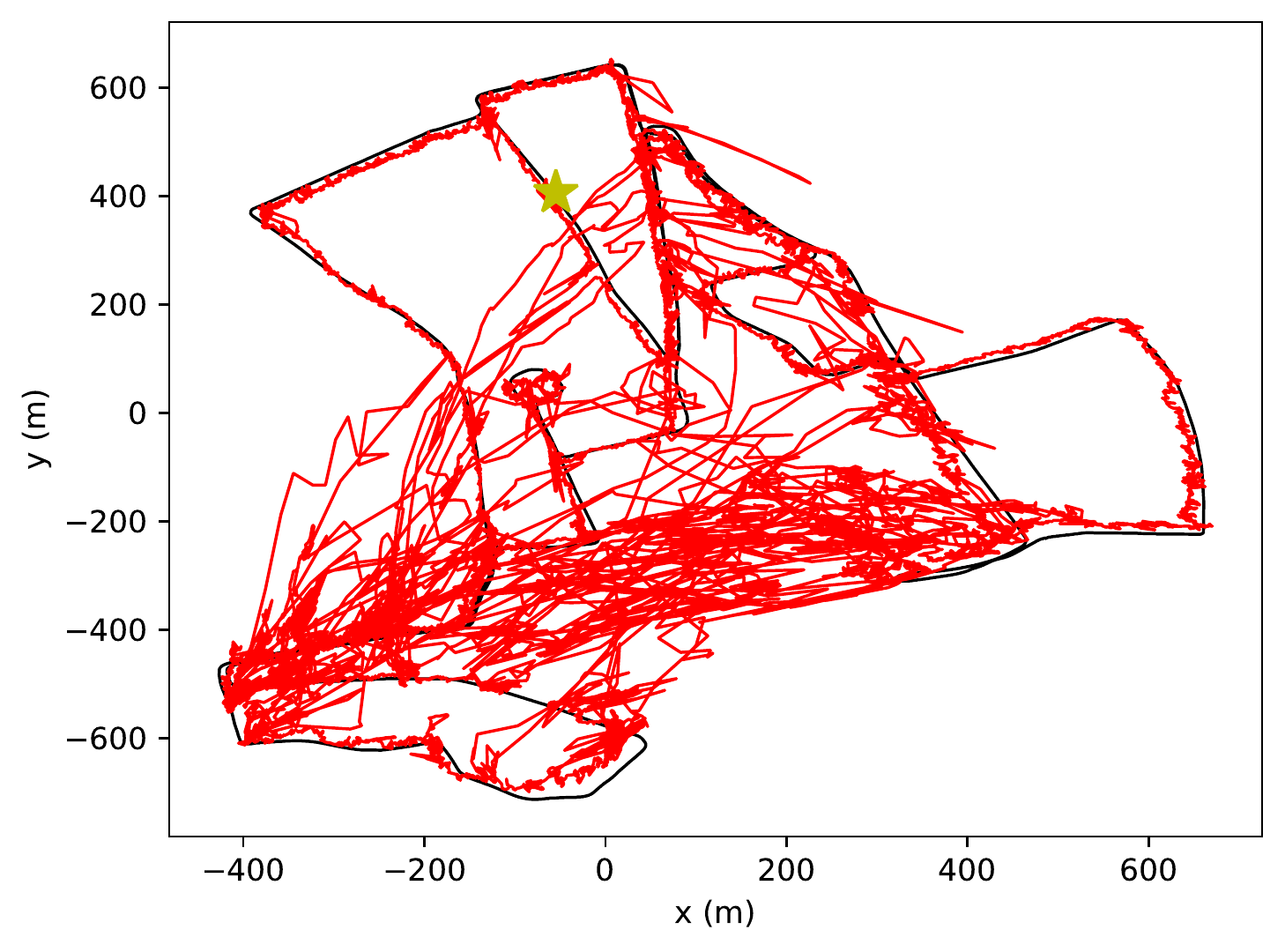}
		MapNet \cite{brahmbhatt2017mapnet}
	\end{minipage}
	\begin{minipage}{0.32\linewidth}
		\centering
		\includegraphics[width=1.0\linewidth]{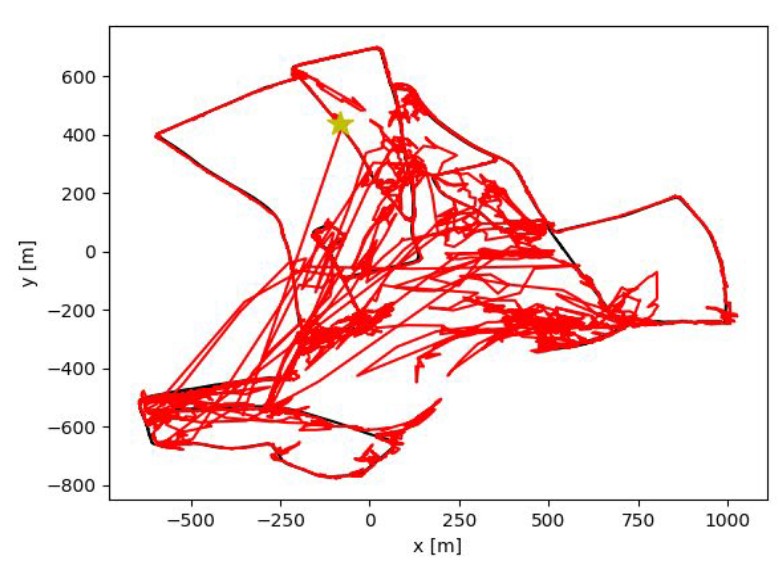}
		AtLoc+ \cite{wang2020atloc}
	\end{minipage}
	\begin{minipage}{0.32\linewidth}
		\centering
		\includegraphics[width=1.0\linewidth]{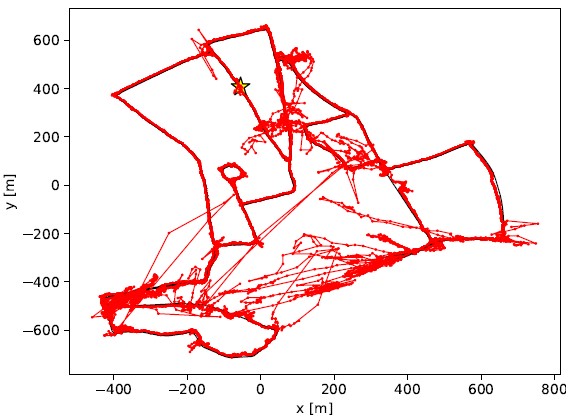}
		GRNet \cite{xue2020learning}
	\end{minipage}
	\begin{minipage}{0.32\linewidth}
		\centering
		\includegraphics[width=1.0\linewidth]{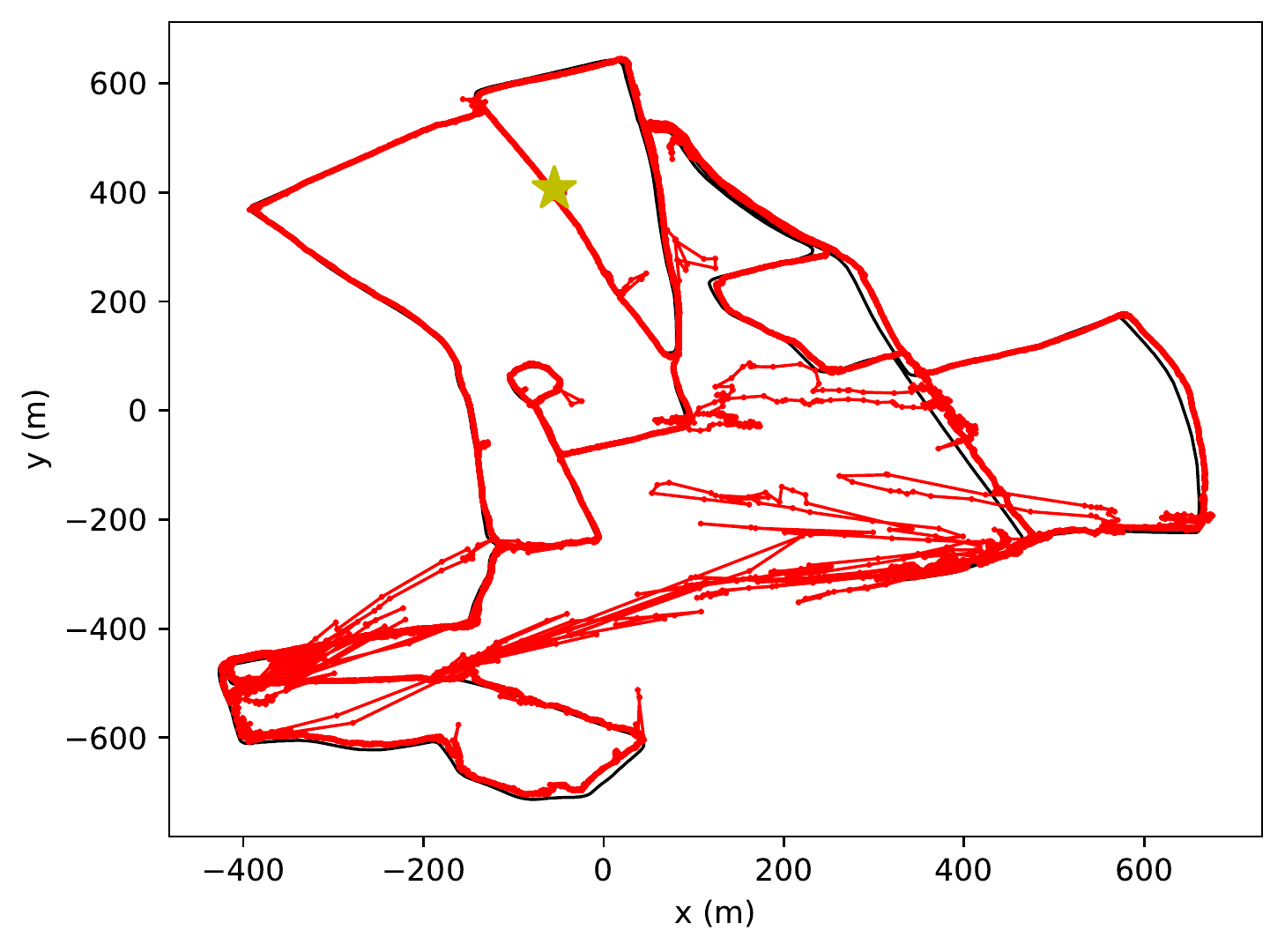}
		ORGMapNet
	\end{minipage}
	\caption{Comparison of trajectories on the FULL1 scene of Oxford RobotCar dataset \cite{maddern20171}. The red and black lines represent the predicted and groundtruth camera poses, respectively.}
	\label{fig:compare_robotcar}
\end{figure}
%\hspace{1mm}

The trajectories obtained by the different models are further illustrated in Fig~\ref{fig:compare_robotcar}. With a single image as input, PoseNet obtains inaccurate estimations with lots of outliers due to scene similarities and over-exposure. In contrast, ORGPoseNet reduces the number of outliers and achieves higher accuracy which is comparable to the sequence based method MapNet. Introducing relative poses between consecutive frames as constraints, MapNet has relatively good pose predictions on the trajectory. However, due to the complexity of the environment, there are still a large number of unstable predictions. In contrast, our proposed ORGMapNet significantly reduces the number of outliers and outperforms the baseline MapNet. This is because we construct the object relation graph on the static objects and enhance the discrimination of image representation by object relation features.

\begin{figure}[t]
	\tiny
	\centering
	\begin{minipage}{0.185\linewidth}
		\centering
		\includegraphics[width=1.0\linewidth]{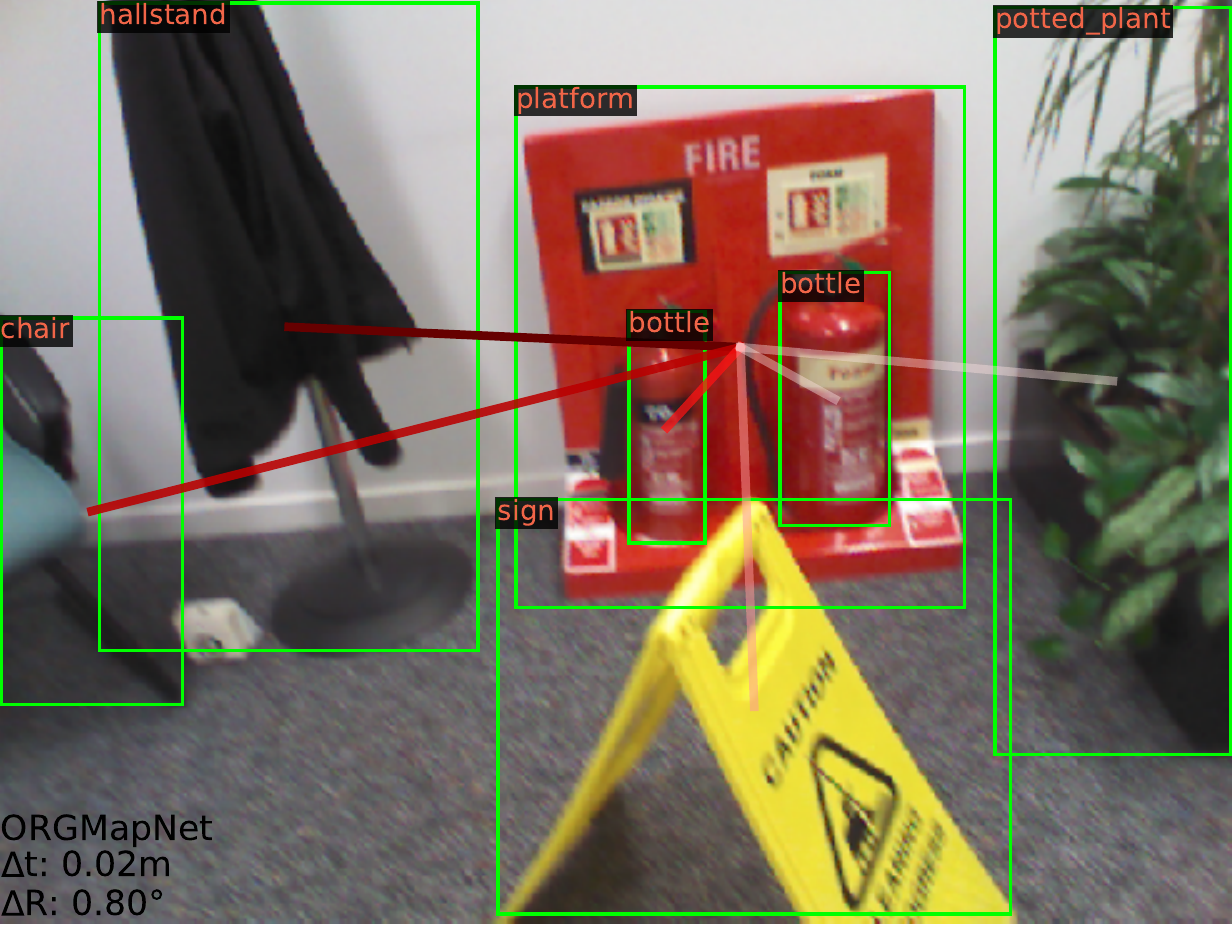}
		7Scenes
	\end{minipage}
	\begin{minipage}{0.185\linewidth}
		\centering
		\includegraphics[width=1.0\linewidth]{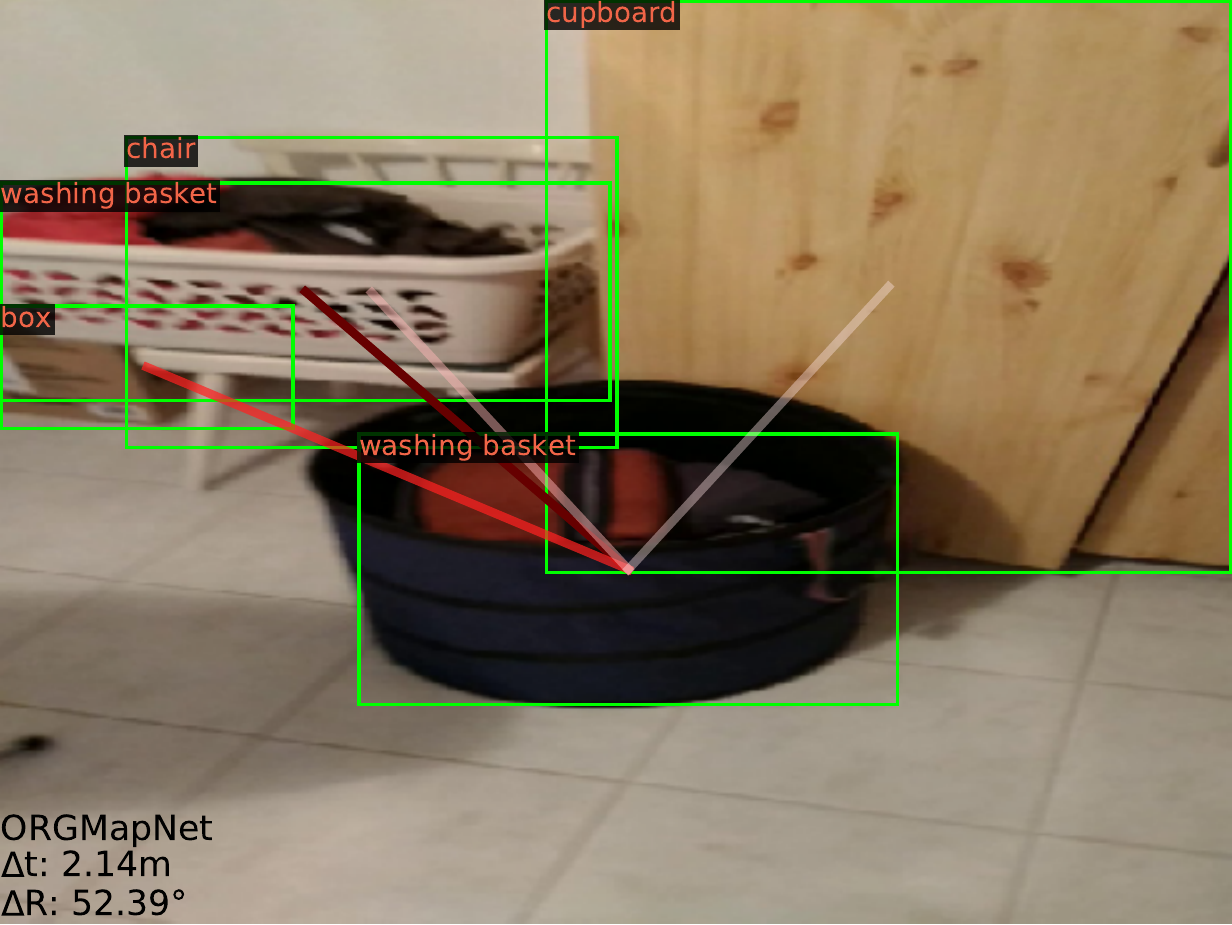}
		RIO10
	\end{minipage}
	\begin{minipage}{0.185\linewidth}
		\centering
		\includegraphics[width=1.0\linewidth]{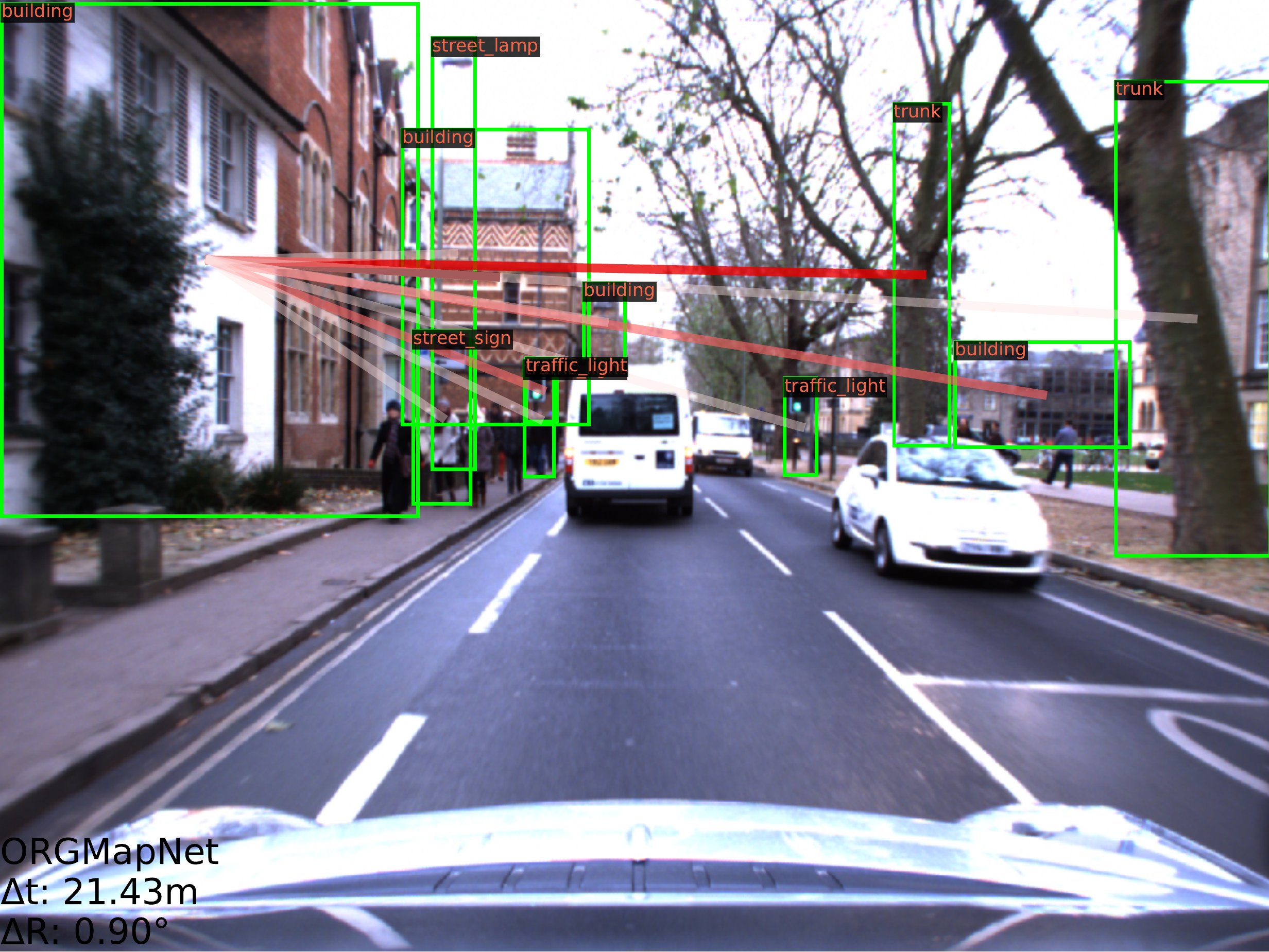}
		RobotCar (overcast)
	\end{minipage}
	\begin{minipage}{0.185\linewidth}
		\centering
		\includegraphics[width=1.0\linewidth]{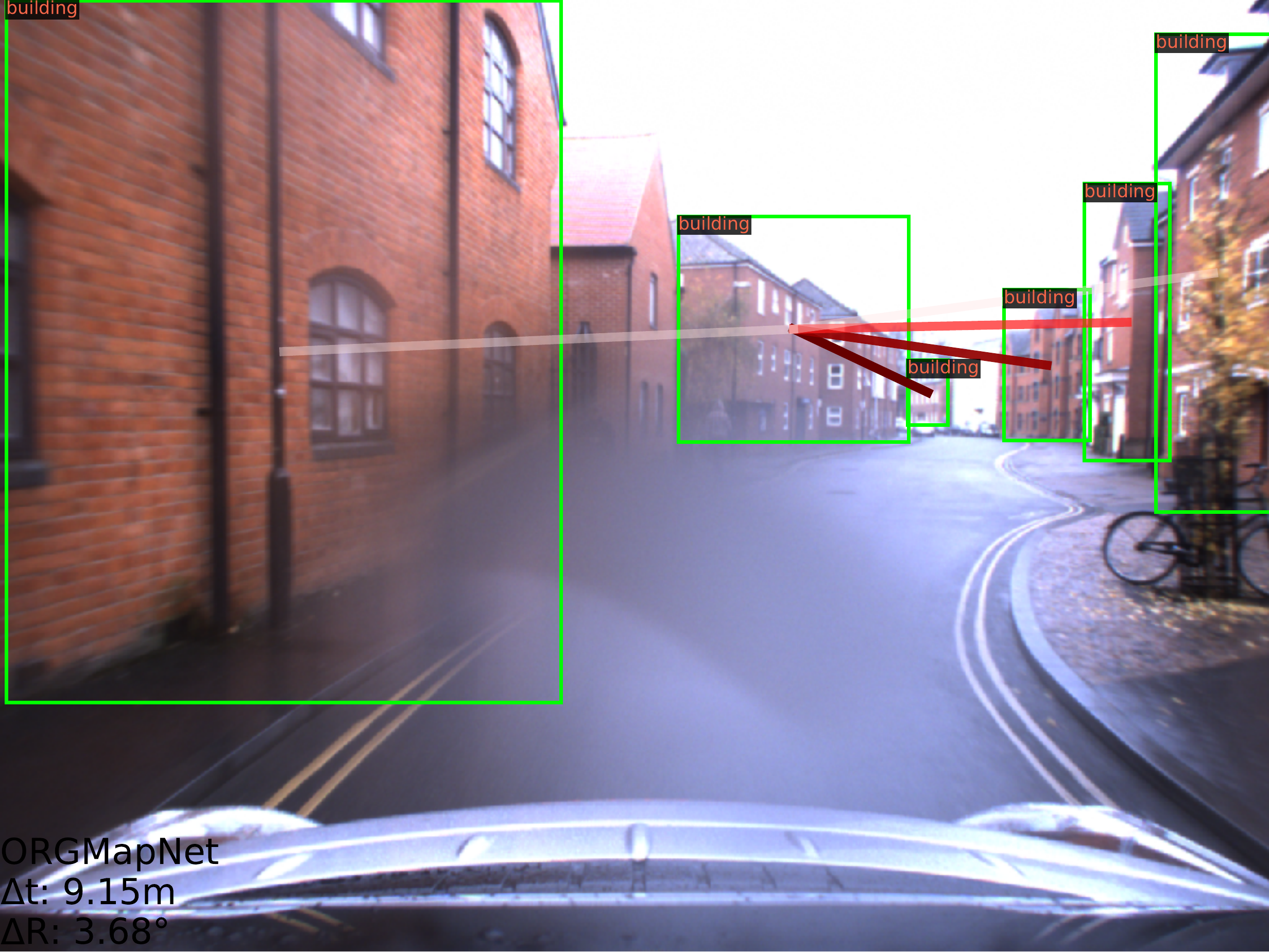}
		RobotCar (rain)
	\end{minipage}
	\begin{minipage}{0.185\linewidth}
		\centering
		\includegraphics[width=1.0\linewidth]{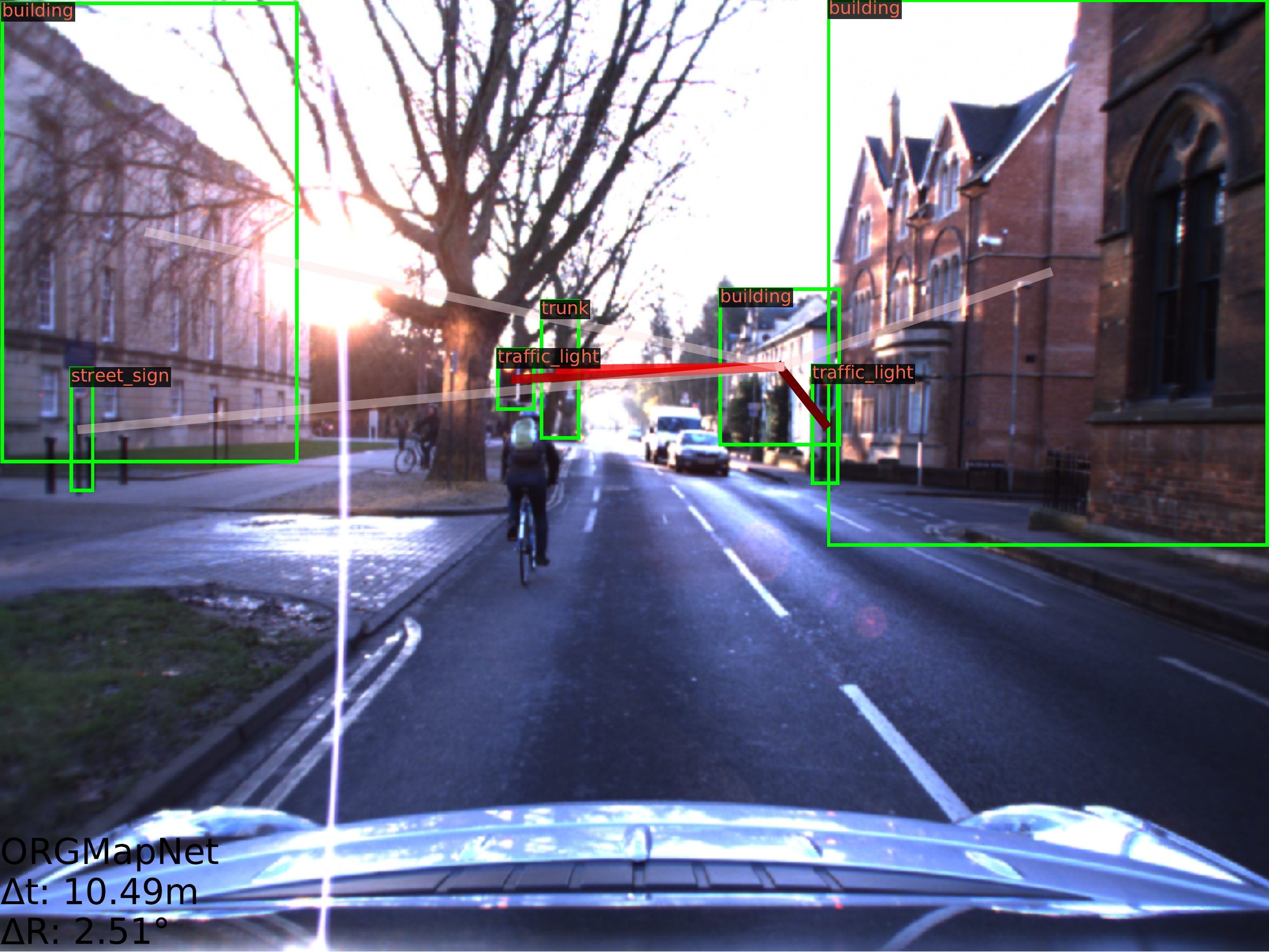}
		RobotCar (sun)
	\end{minipage}
	\begin{minipage}{0.025\linewidth}
		\centering
		\includegraphics[width=1.0\linewidth]{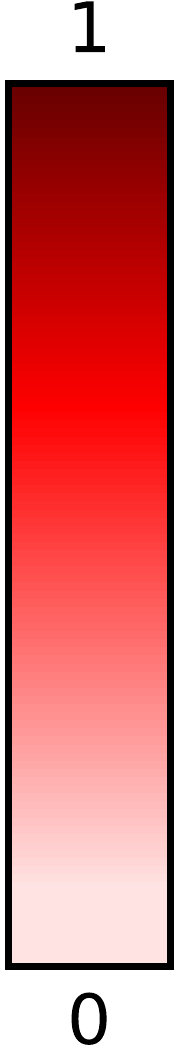}
		%\vspace{0.05mm}
	\end{minipage}
	\caption{Examples of object nodes and the edges in different dataset. For clarity, only the edges of one object node are shown, with the deeper color encodes the smaller distance of objects.}
	\label{fig:object_relation}
\end{figure}

Some examples of the object nodes as well as the edges in different scenes are illustrated in Fig~\ref{fig:object_relation}. Benefiting from the constructed dynamic graph, our ORG can capture the relationships among objects and generate a more discriminative representation for relocalization.

\subsection{Ablation Study}

We conduct ablation studies on RobotCar dataset to verify the configuration of our design. As shown in Table \ref{tab:ablation}, compared to the baseline MapNet \cite{brahmbhatt2017mapnet}, pose errors decrease progressively with the increase of GNN layers. A drastic improvement happens when 3rd layers of GNN is added. Adding more layers can further reduce the error but with more computing cost. Therefore we choose the layer number of GNN as 4. If we reduce the number of nearest neighbor from 5 to 3, or change the aggregate mode of each layer from Max to Sum, the pose error will both increase accordingly. We also test the effectiveness of our dynamic graph update strategy by replacing it with the fixed edge connection from the first layer, the degraded performance verify the necessity of this strategy. Finally, to verify the robustness of the learned object feature with respect to the partial detecting capability, we manually drop out a number of the detected objects with a keeping ratio and only use the rest as input to the GNN. The performance increases with the higher keeping ratio. Surprisingly, keeping only 60\% of the detected objects can achieve good performance close to using all of the objects, demonstrating the robustness of our method to the object detection capability.

\begin{table}[t]
	\small
	%	\footnotesize
	\centering
	\renewcommand{\arraystretch}{0.8}
	\setlength\tabcolsep{8pt}
	\resizebox{1\linewidth}{!}{
		\begin{tabular}{l|l|cc}
			\toprule
			& \multirow{2}{*}{Configuration} & \multicolumn{2}{c}{Scene} \\
			& & FULL1 & FULL2 \\
			\midrule
			\multirow{5}{*}{GNN Layers} & 0 (MapNet)             & 18.21m, 6.66\degree & 21.16m, 6.38\degree \\
			& 1                                                  & 17.14m, 3.26\degree & 17.11m, 2.80\degree \\
			%		\midrule
			& 2                                                  & 15.17m, 2.36\degree & 17.14m, 2.66\degree \\
			%		\midrule
			& 3                                                  & 9.92m, 1.25\degree & 9.28m, 2.36\degree \\
			%		\midrule
			& 4 \textbf{(ORGMapNet)}                             & \textbf{8.84m}, \textbf{1.13\degree} & \textbf{8.01m}, \textbf{1.26\degree} \\
			\midrule
			KNN Neighbors & $k=5\rightarrow3$                    & 10.44m, 1.35\degree & 12.50m, 1.41\degree \\
			\midrule
			Aggregate Mode & max$\rightarrow$sum                 & 11.76m, 1.55\degree & 12.50m, 1.61\degree \\
			\midrule
			Update Mode & dynamic$\rightarrow$static             & 9.93m, 1.54\degree & 10.22m, 1.80\degree \\
			\midrule
			\multirow{5}{*}{Objects Detected} &Keep ratio $=0.2$ & 14.89m, 4.59\degree & 17.76m, 4.65\degree \\
			& Keep ratio $=0.6$                                  & 9.92m, 2.08\degree & 8.66m, 2.05\degree \\
			& Keep ratio $=0.8$                                  & 9.05m, 1.46\degree & 8.29m, 1.62\degree \\
			& Keep ratio $=1$ \textbf{(ORGMapNet)}               & \textbf{8.84m}, \textbf{1.13\degree} & \textbf{8.01m}, \textbf{1.26\degree} \\
			\bottomrule
		\end{tabular}
	}
	\caption{\textbf{Ablation study} on RobotCar dataset. The median pose errors are reported with the best results highlighted.}
	\label{tab:ablation}
\end{table}

\section{Conclusions}

In this paper, we propose to improve the camera relocalization by fusing the object relation feature in the image. We construct an ORG module for the image, and capture the object relation feature by dynamic GNNs. Comparing with the common convolutional features, our object relation feature leverages the semantic stability and the pose awareness which are highly desirable for the relocalization task. We integrate our ORG module to two popular baseline models to verify the effectiveness of idea. Extensive experiments on the public 7-Scenes, RIO10 and Oxford RobotCar datasets are conducted and the results demonstrate that our method is able to significantly improve the relocalization accuracy compared with baselines and outperforms previous approaches. We hope that our work can motivate further interest and study on explicitly introducing object information to learn more robust and discriminative features for camera poses.

\clearpage
{\small
\bibliographystyle{ieee}
\bibliography{egbib}
}

\end{document}